\documentclass{article}
\usepackage{acra}
\usepackage{graphicx}
\usepackage{amsmath}
\usepackage{booktabs} 
\usepackage{multirow} 
\usepackage{array} 
\usepackage{subcaption}
\usepackage{url}
\usepackage{amssymb}
\usepackage{diagbox}
\usepackage{cite}

\usepackage[table]{xcolor}
\usepackage[colorinlistoftodos]{todonotes}
\usepackage[colorlinks=true, allcolors=blue]{hyperref}
\usepackage{acronym}
\usepackage{widetable}
\usepackage{titlesec}
\usepackage{caption}

\definecolor{lightgreen}{RGB}{220,255,220} 
\definecolor{lightblue}{RGB}{220,240,255}  
\definecolor{lightyellow}{RGB}{255,255,220} 
\definecolor{lightgray}{RGB}{240,240,240}  
\definecolor{lightred}{RGB}{255,220,220} 

\title{M2S-RoAD: Multi-Modal Semantic Segmentation for Road Damage Using Camera and LiDAR Data}
\author{Tzu-Yun Tseng$^1$, Hongyu Lyu$^1$, Josephine Li$^2$, Julie Stephany Berrio$^1$,\\ \textbf{Mao Shan$^1$, Stewart Worrall$^1$ }\\ 
$^1$Australian Centre for Robotics (ACFR), The University of Sydney, Australia  \thanks{This research is funded by Transport for New South Wales (TfNSW), iMOVE Australia and supported by the Cooperative Research Centres program, an Australian Government initiative. This research was partially supported by the Australian Government through the Australian Research Council's ARC Training Centre funding scheme for Automated Vehicles in Rural and Remote Regions  (project IC230100001).}\\ $^2$ University of Washington, United States of America \\ $^1$
 \{t.tseng, h.lyu, j.berrio, m.shan, s.worrall\}@acfr.usyd.edu.au, $^2$ josli@uw.edu
}

\begin{document}

\maketitle

\begin{abstract}
Road damage can create safety and comfort challenges for both human drivers and autonomous vehicles (AVs). This damage is particularly prevalent in rural areas due to less frequent surveying and maintenance of roads. Automated detection of pavement deterioration can be used as an input to AVs and driver assistance systems to improve road safety. Current research in this field has predominantly focused on urban environments driven largely by public datasets, while rural areas have received significantly less attention. This paper introduces M2S-RoAD, a dataset for the semantic segmentation of different classes of road damage. M2S-RoAD was collected in various towns across New South Wales, Australia, and labelled for semantic segmentation to identify nine distinct types of road damage. This dataset will be released upon the acceptance of the paper. 
\end{abstract}

\section{Introduction}

Advancements in vehicular technology and artificial intelligence (AI) have led to different use cases where AI is applied to assist drivers or as a module in highly automated vehicles (AVs). This applies to scene understanding, that uses perception sensors like cameras and/or LiDARs to infer the context to the driving environments \cite{berrio2021camera}.
Machine learning, a subset of AI, has proven to reach an impressive level of accuracy at perception tasks. Nevertheless, it requires large amounts of data to train models capable of interpreting the environment around the vehicle \cite{9356353}. While there are many publicly available datasets, they are primarily focused on urban environments. 

\begin{figure}[t!]
    \centering
    
    \begin{subfigure}[t]{.98\columnwidth}   
        \centering 
        \includegraphics[trim={0cm 0cm 0cm 18cm},clip, width=\columnwidth]{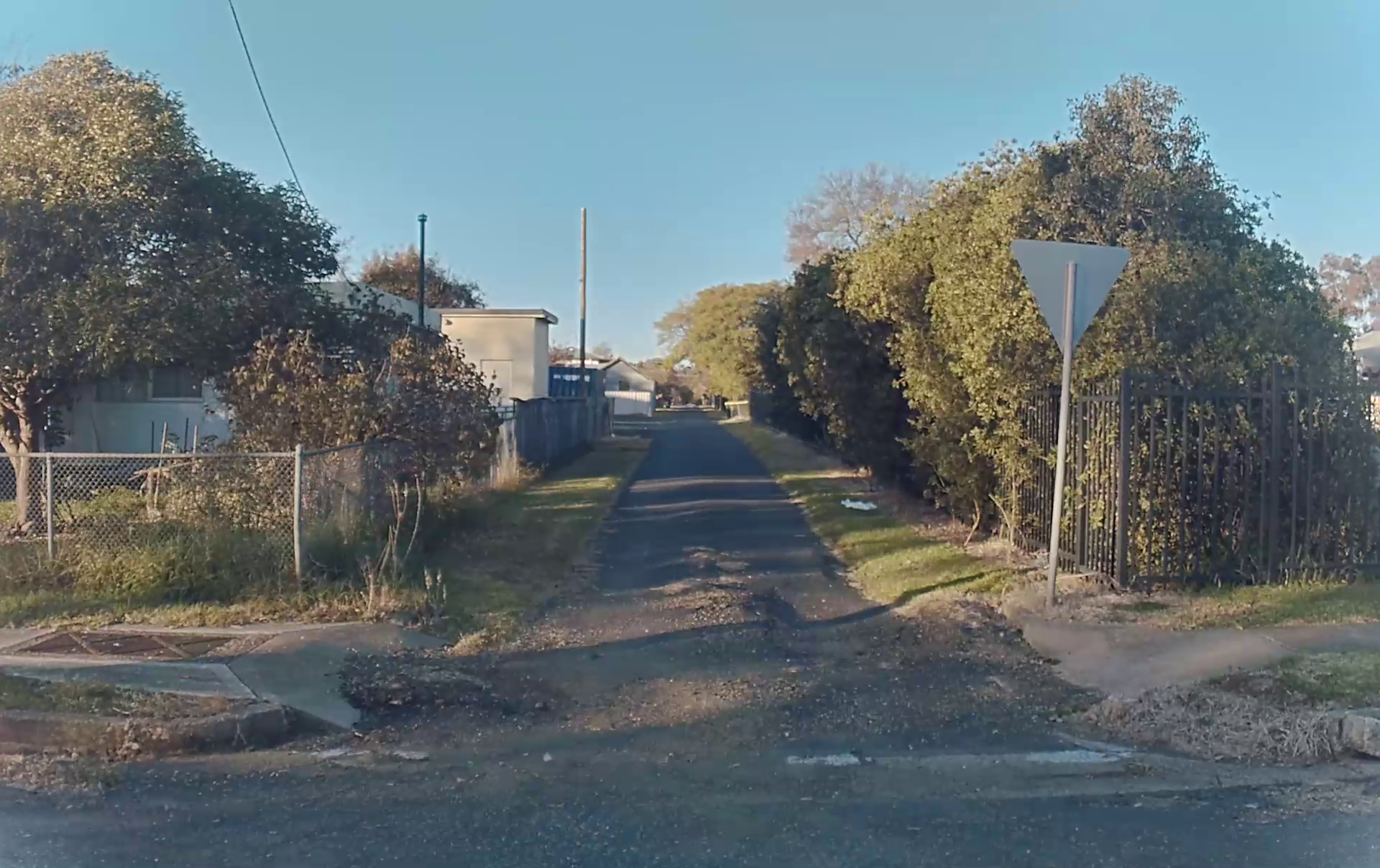}%
        \label{fig:image3}
        \caption{\small Rural road with uneven surface}
    \end{subfigure}
  
    \begin{subfigure}[t]{.98\columnwidth}   
        \centering 
        \includegraphics[trim={0cm 0cm 0cm 18cm},clip, width=\columnwidth]{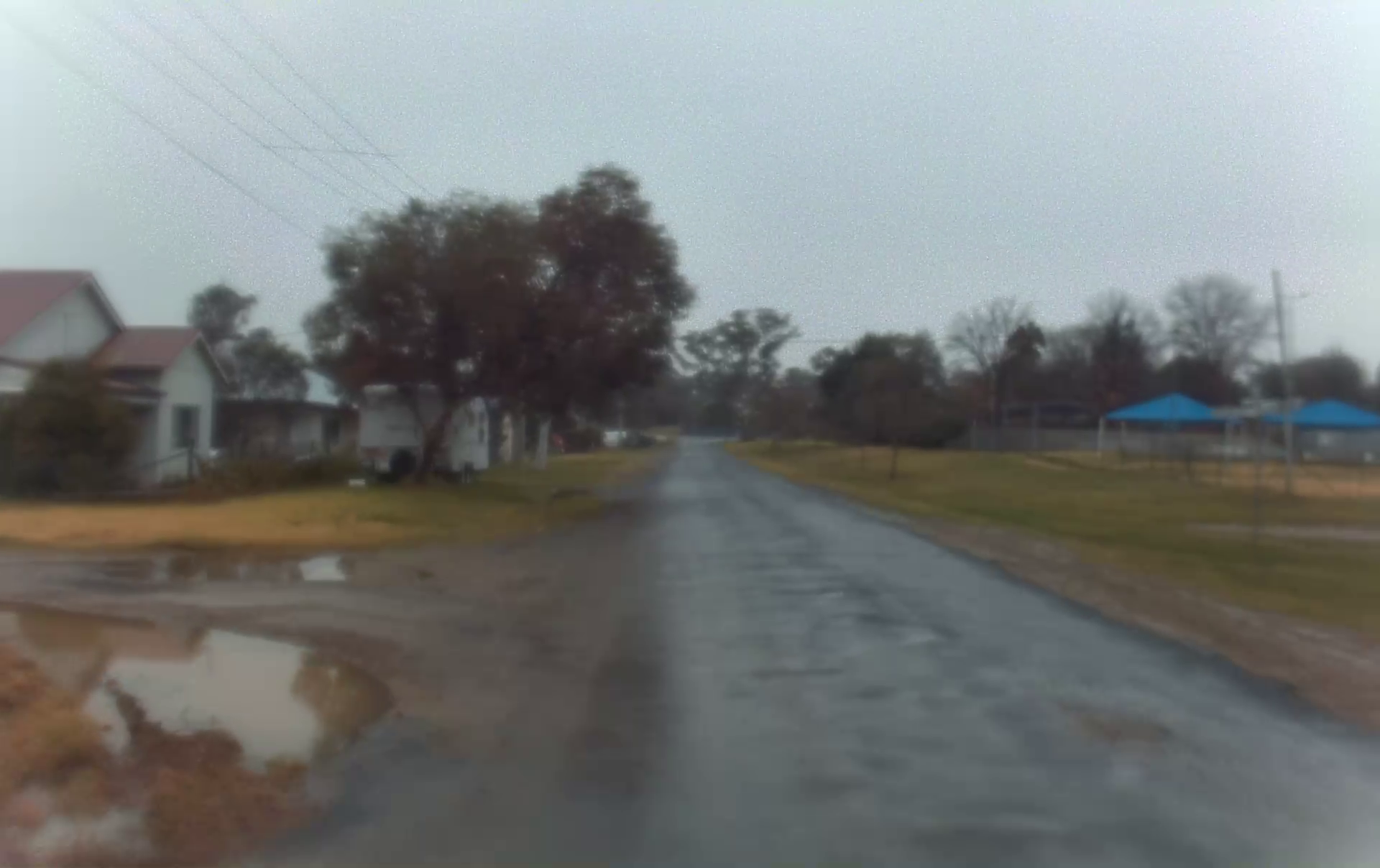}%
        \label{fig:image4}
        \caption{\small Rural road with potholes filled with water}
    \end{subfigure}

    \caption[The images in a 2x2 grid]
    {\small Scenes from rural regions with road damage.} 
    \label{fig:imagegrid}
\end{figure}

One of the objectives of scene understanding is to warn the driver or system when a hazard is detected on the road. Road damage represents a safety risk for all users \cite{shen2015serious}, and local councils may be held liable if an accident occurs. Thus, significant efforts have been made to detect and repair road damage in metropolitan areas \cite{maeda2018road}, but rural regions still lack the infrastructure to quickly address issues such as the ones shown in Fig. \ref{fig:imagegrid}. 

For urban road damage detection, image-based methods are widely used as camera images offer detailed, colour-rich information about the road surface \cite{cao2020survey}. However, camera-only methods are vulnerable to lighting and weather variations, which can significantly impact the accuracy of road damage segmentation. Recently, 3D information, such as disparity maps \cite{li2021road} or LiDAR point clouds \cite{meivel2024remote}, has been incorporated to address this issue. Integrating depth data into these methods has proven to enhance the robustness of road anomaly detection \cite{caltagirone2019lidar}.

For road damage detection, there is a lack in publicly available datasets that include LiDAR point clouds, making it challenging to evaluate and improve methodologies proposed in the literature \cite{doi:10.1061/40794(179)103}. Due to this, it is also difficult to innovate and create new methods for identifying road damage. 

This proposed dataset aims to address the gaps in existing datasets for road damage, particularly the underrepresentation of rural environments and the absence of datasets that include LiDAR point clouds. The M2D-RoAD dataset provides camera images, LiDAR point clouds, and sensor calibration, enabling research on multi-modal segmentation of road damage. 

The contributions of this paper are listed as follow:

\begin{itemize}
    \item M2S-RoAD is the first dataset for semantic segmentation of road damage collected under different weather and illumination conditions in several cities and road scenes. The annotations cover 9 types of road damage.  
    \item M2S-RoAD is the first multi-modal dataset to fuse polygon-labelled images with corresponding point clouds to accurately understand the shape and depth of the road damage.
    \item  Benchmark evaluation and analysis on five different image-based semantic segmentation techniques as a baseline for comparisons on new methods.
\end{itemize}

\section{Related Datasets}

Multiple datasets have been developed for detecting road surface damage, with most relying on single images, and some incorporating stereo vision to estimate depth as shown in Table \ref{tab:datasets}. Image-based approaches to object identification generally fall into two categories: detection using 2D bounding boxes and pixel-level semantic segmentation. This section reviews the datasets available in the literature, focusing on the sensors used and the types of annotations provided.

\subsection{RGB Image-Based Datasets}

The two main types of annotations used in RGB image-based datasets are pixel-wise semantic labels and 2D bounding boxes.

\subsubsection{Pixel-Level Annotated Datasets}

\textit{Cracktree206 Dataset:} Cracktree206 \cite{zou2012cracktree} provides 206 images of road surfaces captured using an area-array camera. The dataset includes pixel-level annotations and augmented images. Each image contains manually annotated ground-truth crack curves.

\textit{CCSAD Dataset:} The CCSAD dataset \cite{guzman2015towards} was collected in Mexico with 500 GB of stereo images, videos, and metadata. The dataset is divided into four parts: Colonial Town Streets, Urban Streets, Avenues, and Small Roads and Tunnel Networks. It provides depth and stereo data which are highly affected by lighting and shadow conditions.

\textit{CrackForest Dataset:}  The CrackForest dataset \cite{shi2016automatic} consists of 118 high-resolution images of cracks took by iPhone 5 in urban Beijing. Each image has been resized to 480×320 pixels for uniformity with pixel-wise annotations.

\textit{CRACK500 Dataset:} CRACK500 \cite{zhang2016road} is aiming specifically to detect crack. Composed of 1,896 training images, 328 validation images, and 1,124 test images, and each image is further divided into smaller cropped parts to increase the dataset size. Pixel-wise annotations by multiple annotators are provided for cracks.


\textit{Deepcrack Dataset:} The DeepCrack dataset \cite{liu2019deepcrack} consists of 537 RGB colour images with manually annotated segments. Each image has a resolution of 544 × 384 pixels. The dataset is divided into two subsets: a training set with 300 images and a testing set with 237 images. Pixel-level annotations are provided

\textit{NHA12D Dataset:} The NHA12D dataset \cite{huang2022nha12d} includes 80 images of concrete and asphalt pavements, collected by vehicles on the A12 network in the UK. The images have a resolution of 1928×1080 and pixel-wise annotations for crack and non-crack regions. The dataset captures a variety of road features, including road markings, vehicles, water stains, road studs, and other random objects.

\textit{CQU-BPDD Dataset:} The CQU-BPDD dataset \cite{tang2021iteratively} consists of 60,059 high-resolution asphalt road images (1200×900 pixels), collected in southern China using in-vehicle cameras. Also provided with patch-level annotations for seven types of pavement damage, including transverse cracks, longitudinal cracks, alligator cracks, crack pouring, raveling, repairs, and normal conditions. The CQU-BPDD dataset captures 2×3 meter pavement patches, each image regarded as a candidate box for binary image classification.

\textit{EdmCrack1000 Dataset:} EdmCrack1000 \cite{mei2020densely} contains 1000 images, each with a resolution of 1920×1080 pixels for detecting cracks. The camera is installed outside and rear of the car to avoid interference from windows or the windshield.

\begin{table*}[ht]
\centering
\caption{\small Comparison of road damage datasets.  Pixel-wise label (PL) datasets often fall into the binary classification (BC) category to differentiate between crack and background. As for datasets with bounding-box (BB) label, even though they are Multi Class (MC) they are not as accurate as compared to those with PL-based labels.} 
\label{tab:datasets}

\resizebox{\textwidth}{!}{%
\begin{tabular}{|m{3cm}|m{2.5cm}|m{1cm}|m{2.5cm}|m{3cm}|m{2cm}|m{5cm}|m{1cm}|}
\hline
\textbf{Dataset} & \textbf{Camera} & \textbf{Label} & \textbf{Resolution} & \textbf{Size} & \textbf{Location} & \textbf{Features} & \textbf{Class} \\ \hline
Cracktree206 & Area-array & PL & 800×600 & 206 images & N/A & Augmented by rotation and flipping & BC \\ \hline
CCSAD & scA1300-32fm & PL & 1096×822 & 500 GB & Mexico & Lighting, reflection, and shadow & MC \\ \hline
CrackForest & iPhone 5 & PL & 480×320 & 118 images & Beijing & Urban road cracks & BC \\ \hline
CRACK500 & Smartphone & PL & Varied & 3.3k images & Multiple & Dataset divided into cropped parts & BC \\ \hline
Malillary Vistas & Varied & PL, IL & 1920×1080 & 25k images & Worldwide & World-wide coverage & MC \\ \hline
Deepcrack & Line-array & PL & 544×384 & 537 images & N/A & Asphalt and concrete road & BC \\ \hline
NHA12D & Camera & PL & 1920×1080 & 80 images & UK & Concrete and asphalt pavements & BC \\ \hline
CQU-BPDD & Vehicle-mounted & Patch-level & 1200×900 & 60k images & China & Candidate box & BC \\ \hline
EdmCrack1000 & Vehicle-mounted & PL & 1920×1080 & 1k images & Canada & Shadows and occlusions & BC \\ \hline
RDD2018 & Smartphone & BB & 600×600 & 9k images & Japan & Eight types of damage & MC \\ \hline
RDD2019 & Smartphone & BB & 128×128 & 13.1k images & Japan & Night types of damage & MC \\ \hline
GAPs & Monochrome & BB & 1920×1080 & 2k grayscale images & Germany & Patch size 64×64 & MC \\ \hline
RDD2020 & Smartphone & BB & 600×600 to 720×720 & 26.3k images & India, Japan, Czech Republic & Diverse road conditions & MC \\ \hline
RDD2022 & Smartphone, drone, motorbike & BB & 512×512 to 3650×2044 & 47.4k images & 6 countries & Expanded RDD2020 to 3 more countries & MC \\ \hline
SHREC 2022 & RGB-D & PL & Varied & 4.3k image/mask pairs, 797 video clips & Multiple & Potholes and cracks & BC \\ \hline
Pothole-600 & ZED Stereo & PL & Varied & 67 stereo image pairs & N/A & Disparity transform & MC \\ \hline
Sunny & GoPro & N/A & 3680×2760 & 48.9k images & N/A & Image processing & N/A \\ \hline
M2S-RoAD & Vehicle-mounted & PL & 1928×1208 & 1071 images and 43911  point clouds& NSW, Australia & Different weather and illumination conditions  & MC \\ \hline

\end{tabular}%
}

\end{table*}

\subsubsection{Bounding-Box Annotated Datasets}

\textit{RDD2018 Dataset:} The RDD2018 dataset \cite{maeda2018road} contains 9,053 road images (600 × 600 pixels) captured across seven Japanese cities, with 15,435 annotated bounding boxes. It spans a range of weather and lighting with the bounding boxes annotation with smartphone shoot 10m ahead.

\textit{RDD2019 Dataset:} RDD2019 \cite{maeda2021generative} further enrich the RDD2018 with utility hole as a new class and resize into 128X128 pixels. RDD2019 datasets consists of 13,135 images augmented by GAN and 30,989 annotated bounding boxes but shared the same coverage as RDD2018 in Japan.

\textit{German Asphalt Pavement Distress (GAPs) Dataset:} The GAPs dataset \cite{eisenbach2017get} consists of 1,969 gray-scale images with bounding-box annotations for various types of road damage, including cracks, potholes, inlaid patches, applied patches and open joints. Each image has a resolution of 192 ×1080 pixels. The dataset is divided into 1,418 training images, 51 validation images, 500 validation-test images and 500 test images. Annotations are by bounding boxes containing detailed damage categories.

\textit{RDD2020 Dataset:} RDD2020 \cite{arya2020global} expand the coverage to the Czech Republic and India to enrich the datasets with 26,336 images. It contains 4 different damage type instances, such as potholes, longitudinal cracks, alligator cracks and lateral cracks, total round to 25,046 instances with two resolutions 600×600 and 720×720 respectively.

\textit{RDD2022 Dataset:} The RDD2022 dataset \cite{arya2022rdd2022} is the most extensive version of the RDD series, with 47,420 images and 55,007 bounding box annotations. The coverage expands to six different countries with the image captured by different sensors including smartphone, camera and google street view from vehicles, drones and motorbikes. The image resolutions varies from 512×512, 600×600 and 720×720 to 3650×2044.

\subsection{Depth Sensor-Based Datasets}

\textit{Sunny Dataset:} The Sunny dataset \cite{nienaber2015detecting} consists of 48,913 high-resolution images (3680×2760 pixels) captured with a GoPro camera mounted in a vehicle. Depth information is provided through stereo matching but does not specify about the labels, as this dataset comes up with image processing rather than machine learning.

\textit{SHREC 2022 Dataset:} The SHREC 2022 RGB-D dataset \cite{thompson2022shrec} includes 4,340 pixel- wise image/mask pairs and 797 unannotated RGB-D video clips, with depth and disparity maps for just pothole and crack detection.  The resolution is 1920×1080 for RGB-D video and is 640×400 for disparity videos at grayscale, as disparity images help to detect road damage.

\textit{Pothole-600 Dataset:} Pothole-600 \cite{fan2019pothole} consists of 67 pairs of stereo images captured using a ZED stereo camera. The dataset provides depth information through disparity maps and the resolutions for different subsets are 1028×1730, 1030×1720 and 1028×1710.

\begin{figure*}[t]
    \centering
    \includegraphics[width=\textwidth]{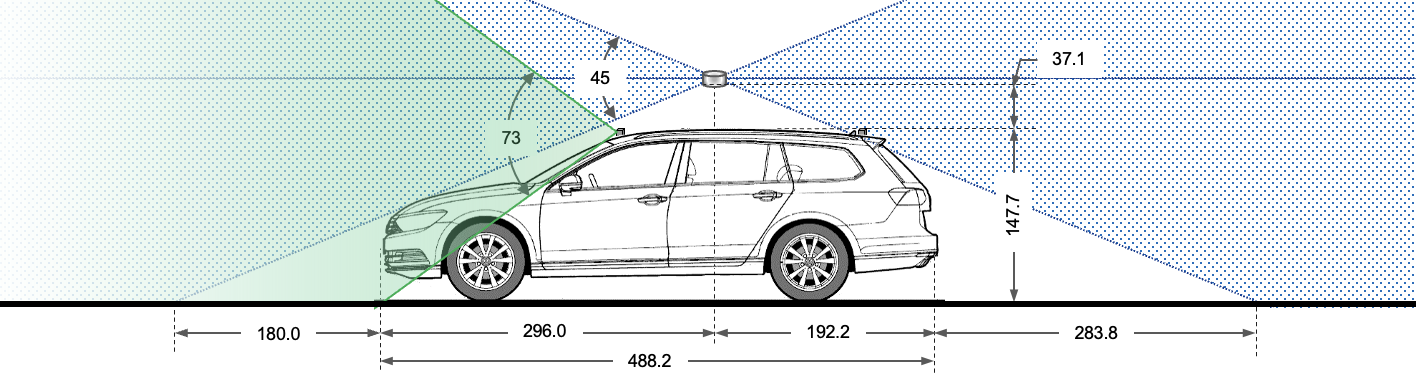}
    \caption{\small Sensor configuration of the data collection platform equipped with a 360-degree LiDAR and a front-facing camera.}
    \label{fig:sensor_setup}
\end{figure*}
\textit{CyberPothole Dataset:} This synthetic dataset \cite{cyberpothole} for pothole detection provides depth maps with 2D bounding boxes annotations. The data was created using templates based on real-world images and rendering the potholes in blender to simulate the depth maps.

This paper presents M2S-RoAD, the first dataset with semantic segmentation labels and corresponding point clouds annotated on 9 types of road damage. The data was collected with a front-facing camera (1928×1208 resolution) and one LiDAR (128 beams). The data logging occurred in rural regions across different weather, illumination, cities, and road types. 


\section{The M2S-RoAD Dataset}

Data collection, post-processing, and annotating the images for semantic segments containing road damage is required to create this comprehensive dataset. This dataset is necessary for training and validating a computer vision system able to detect and classify road damage in diverse and representative environmental conditions. This section outlines the entire process, from sensor setup to creating the final labelled dataset.

\subsection{Platform}

We employed two synchronised sensors for data collection: a front-facing camera and a LiDAR, both mounted on an vehicle (Volkswagen Passat). The SF3324 automotive GMSL model's camera is integrated with an ONSEMI CMOS Image Sensor AR0231 (2M pixels) and a SEKONIX ultra-high-resolution lens. This configuration offers a horizontal field of view (FOV) of 120 degrees and a vertical FOV of 73 degrees, capturing images at a resolution of 1928×1208 pixels (2.3M pixels) \cite{henry2024}.
The LiDAR sensor is an Ouster OS1-128 mounted on the roof. It provides  360-degree 3D data coverage and enables the real-time reconstruction of depth images, signal images, and ambient images of the driving environment. The OS1-128 features a vertical resolution of 128 beams within a 45-degree vertical FOV and a maximum range of 120 meters.

As illustrated in Fig. \ref{fig:sensor_setup}, the camera and LiDAR were mounted horizontally and centrally on the data collection platform at heights of 1.47 meters and 1.84 meters above the ground, respectively.

\subsubsection{Calibration}

\begin{figure}[b!]
    \centering
    \includegraphics[trim={10cm 2cm 10cm 3cm},clip, width=\columnwidth]{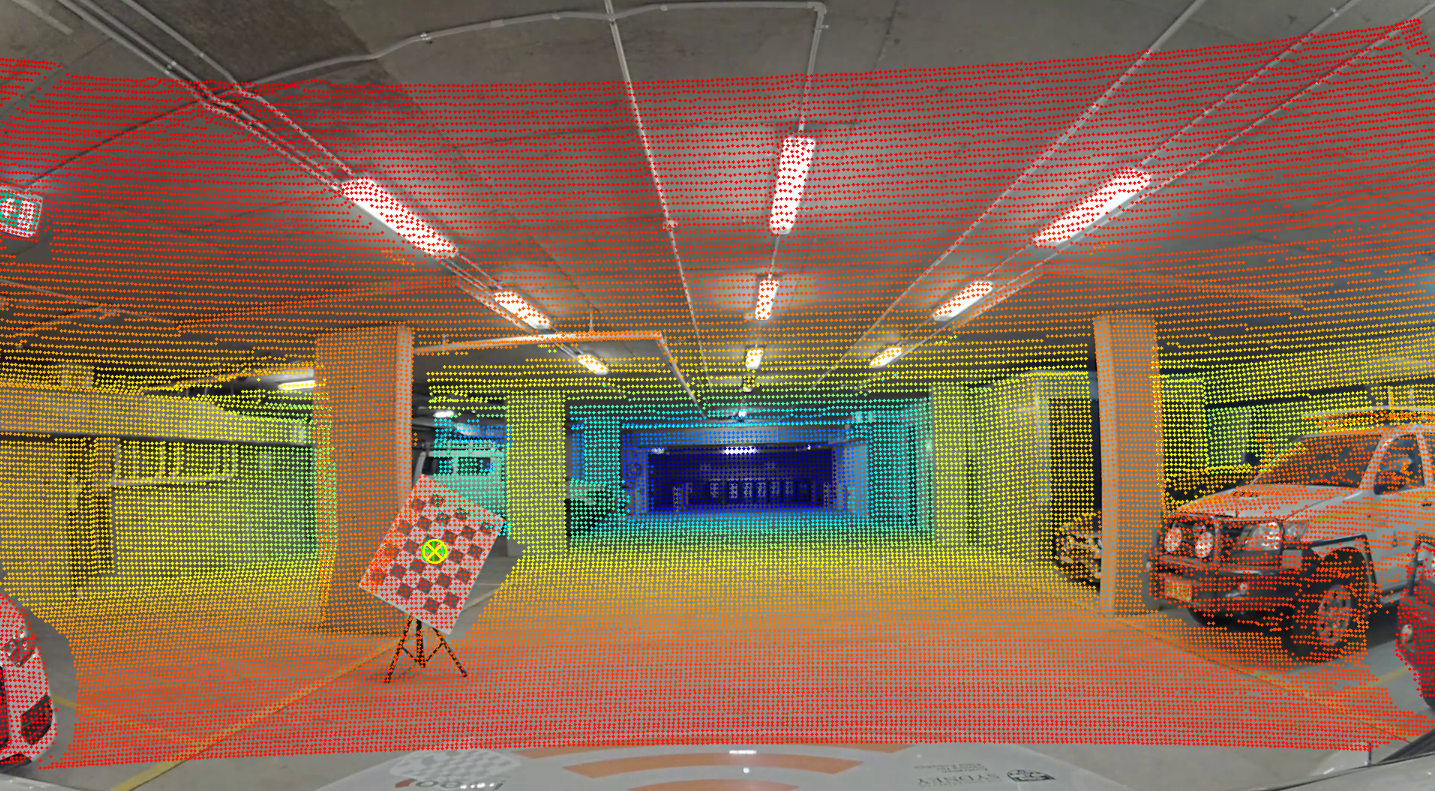}
    \caption{\small LiDAR point cloud projected onto the camera image using the camera's intrinsic parameters and the extrinsic calibration between the camera and LiDAR. }
    \label{fig:Calibration}
\end{figure}
The camera calibration process estimates the lens and image parameters, which enables the correction of lens distortion, accurate measurement of object sizes, and calculation of the camera's position within the scene. Intrinsic calibration is performed using the fisheye camera model \cite{fisheye_model}, with parameters including focal length, principal points, and four fisheye equidistant distortion coefficients. For this calibration, we use the ROS image pipeline \cite{imagepipeline}. The process involves positioning a checkerboard (calibration target) across the entire field of view at close range and in various orientations. An optimisation algorithm is then employed to calculate the camera and image parameters.

Extrinsic sensor calibration specifies the spatial relationship between the camera and LiDAR, allowing for integrating data from multiple sensors into a unified coordinate frame. For this purpose, we employed our lidar-camera calibration package \cite{surabhi_calib_2019} \cite{Darren_2021} to calculate the transformation matrix between the camera and LiDAR frames. These transformations are represented as a 6-degree-of-freedom (6-DOF) pose, encompassing [x, y, z] coordinates and [roll, pitch, yaw] angles.

Fig. \ref{fig:Calibration} shows each point in the LiDAR point cloud overlaid onto the corresponding location in the camera's image, mapping the 3D data from the LiDAR onto the camera's 2D image plane. The colours of the points depict the distance from the camera to the objects in the scene. Red indicates points that are close to the camera, while blue points are farther away.
This process allows for the fusion of data from both sensors.

\subsubsection{Data Collection}

The data was logged while driving across various towns in New South Wales, Australia, including Cudal, Katoomba, Lithgow, and Bathurst. We selected these places to capture a different environmental conditions. The data was collected under varying weather and illumination conditions, including sunny, cloudy, rainy, and overcast conditions.

\begin{figure}[t!]
\centering

\begin{subfigure}[]{\columnwidth}
\centering
\includegraphics[trim={0cm 7cm 0cm 23cm},clip, width=\columnwidth]{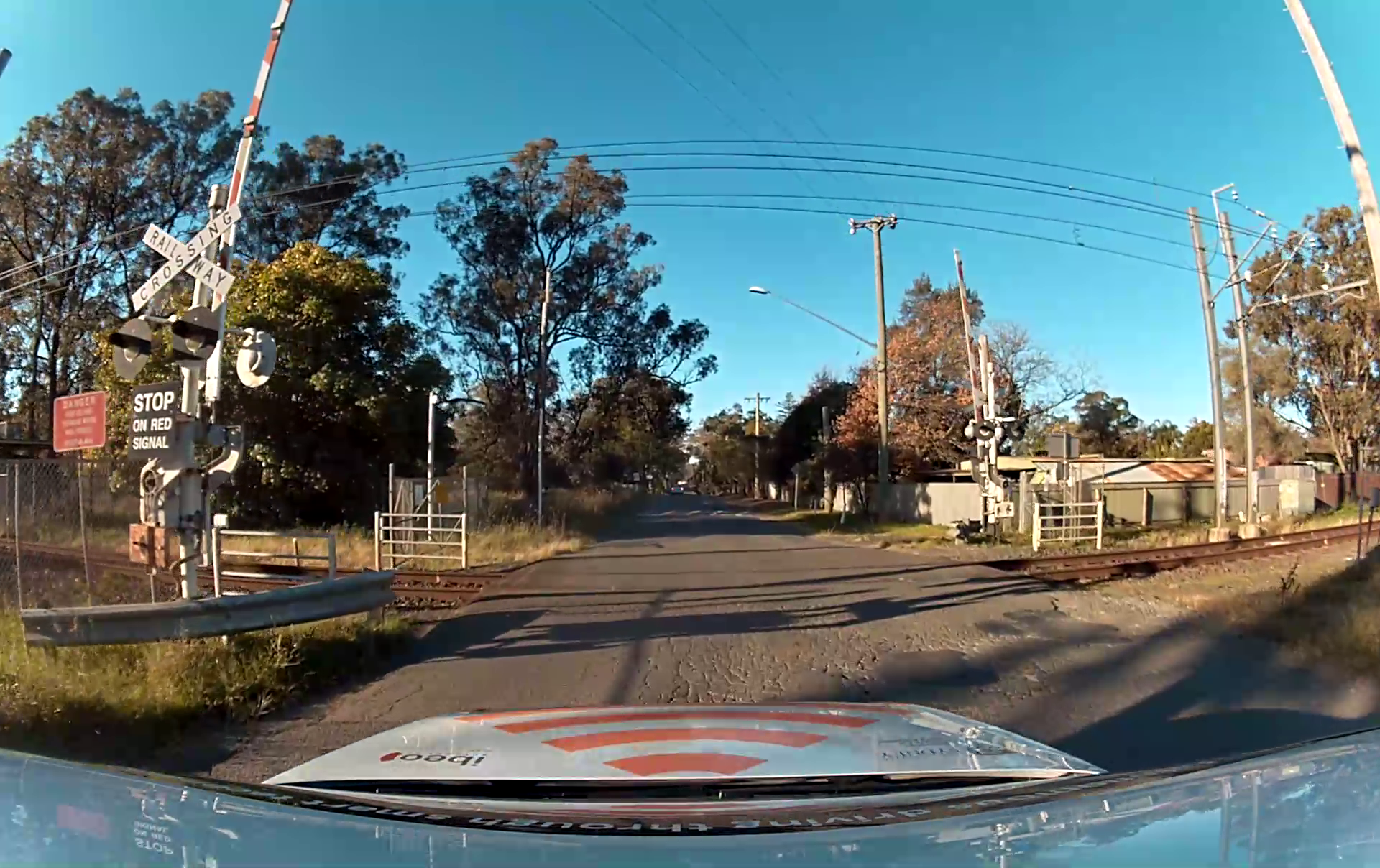}
\caption{Road surface at a road crossing during sunny weather}
\label{fig:instance}
\end{subfigure}

\begin{subfigure}[]{\columnwidth}
\centering
\includegraphics[trim={0cm 7cm 0cm 23cm},clip, width=\columnwidth]{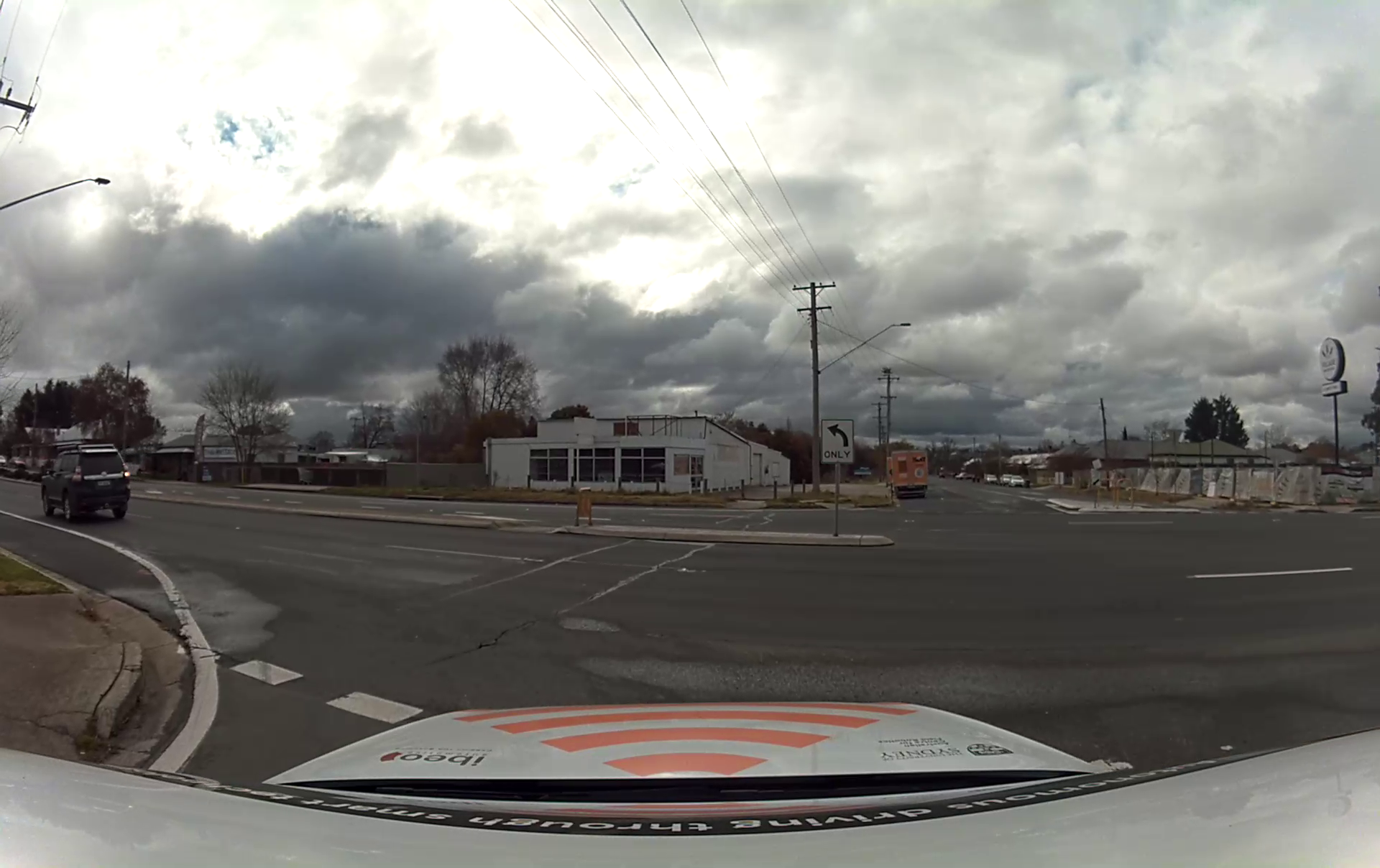}
\caption{Road surface in Bathurst during cloudy weather}
\label{fig:panoptic}
\end{subfigure}

\begin{subfigure}[]{\columnwidth}
\centering
\includegraphics[trim={0cm 7cm 0cm 23cm},clip, width=\columnwidth]{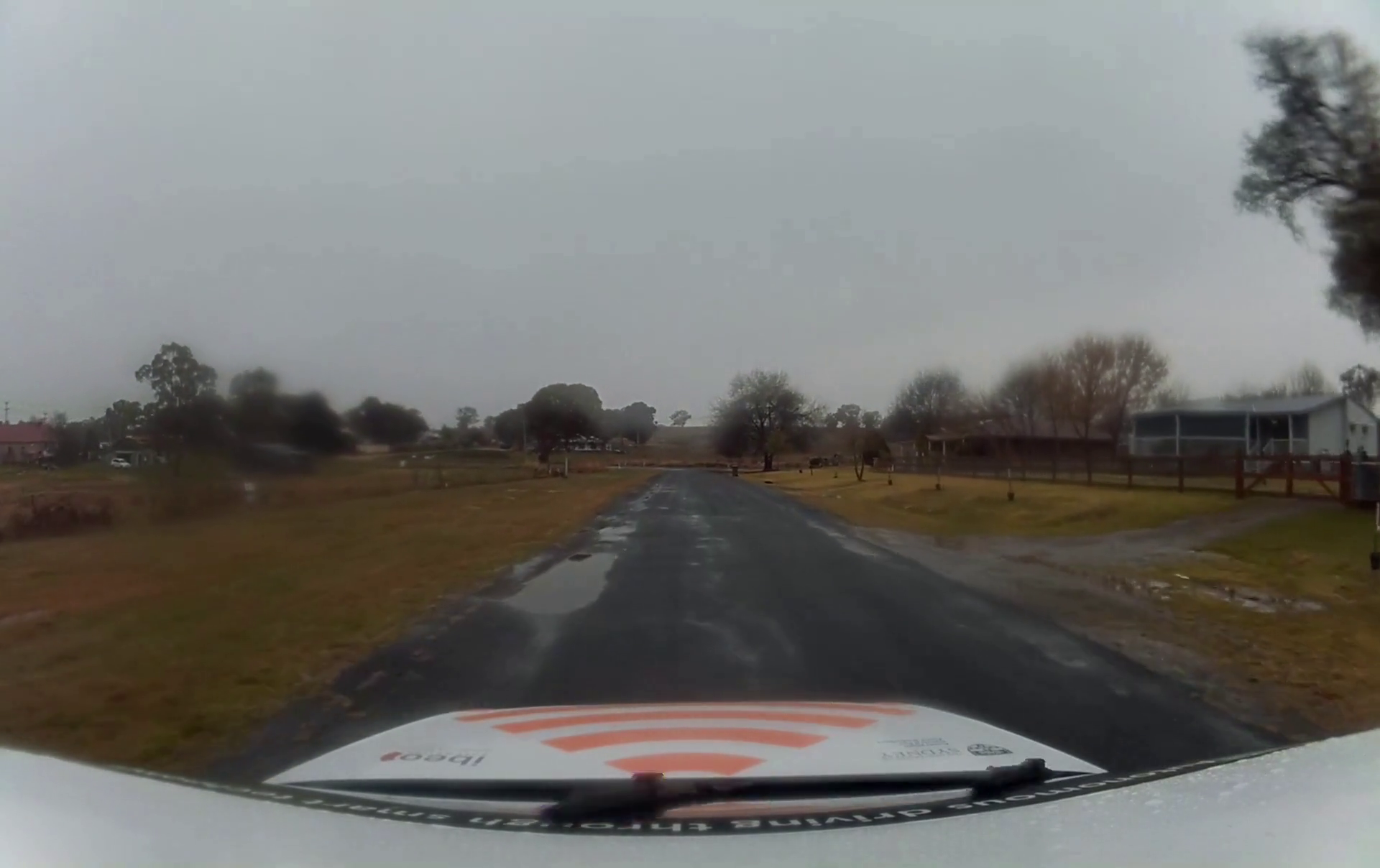}
\caption{Road surface in Cudal during rainy weather}
\label{fig:original}
\end{subfigure}

\begin{subfigure}[]{\columnwidth}
\centering
\includegraphics[trim={0cm 7cm 0cm 23cm},clip, width=\columnwidth]{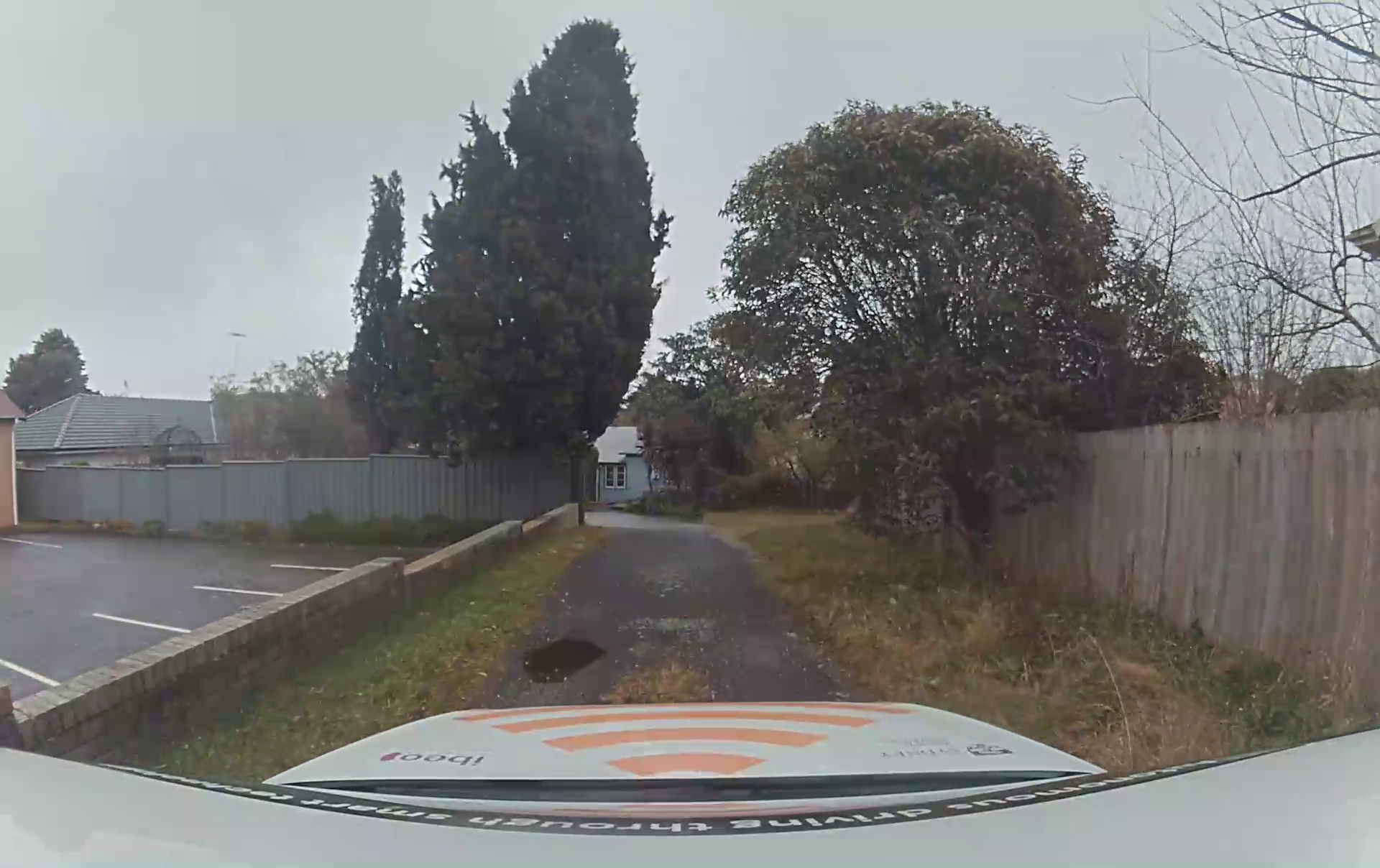}
\caption{Road surface in Katoomba during overcast weather}
\label{fig:semantic}
\end{subfigure}

\caption{\small   Road surface conditions in various regions of New South Wales under different weather conditions.}
\label{fig:images-dataset}
\end{figure}

The images in Fig. \ref{fig:images-dataset} showcase road surface conditions across different regions of New South Wales under varying weather conditions. The first image depicts a wet road in Cudal during rainy weather, highlighting the presence of puddles in a rural setting. The second image shows a narrow road in Katoomba under overcast skies, with dry yet shaded conditions. The third image captures a railway crossing on a sunny day, where the road surface is well-lit and dry. Finally, the fourth image presents an urban road intersection in Bathurst during cloudy weather, with visible road markings and dry pavement. 

Using data taken from a large variety of weather conditions was important due to the way the weather affected the presentation of anomalies. For instance, the depth of a pothole looks very different under sunny weather versus overcast skies; additionally, certain categories such as ``pothole with water" would only show up under certain weather conditions. It was also necessary to take into consideration the variation in visibility caused by abnormal weather conditions; for example, in heavy rain or extremely sunny weather, the ability of the camera to accurately capture the surroundings is decreased.

\subsubsection{Image Selection}

Images were selected to provide a wide variety of examples in terms of various types of road damage as well as across different weather and road conditions. Footage was first selected to encompass different kinds of environments and weather. However, as the images were taken as frames from the video and our original pool comprised of thousands of images, many of the images did not have any road damage, or did not have the categories of damage that we are looking to highlight. The final set of images to be annotated were chosen based on how much value they could contribute to the resulting dataset. We selected images that were spread out throughout the videos and showed anomalies in a range of different environments, such as in different positions on the road, having different degrees of severity, or under varying levels of visibility. Furthermore, a conscious effort was made to balance the amount of images that showcased each category of damages; images that contained significantly more common categories were sometimes passed over in order to minimise the difference between the number of images in each category.

All of the images were categorised into four subsets to reflect the weather conditions during data collection. The subsets include: \textit{Sunny}, \textit{Cloudy}, \textit{Rainy} and \textit{Overcast} conditions.

Since the camera and LiDAR are synchronised, we use the camera-image timestamp to extract 20 preceding and 20 following LiDAR point clouds (captured at a rate of 10 Hz). This approach allows us to combine the point clouds, creating a denser 3D representation that simplifies the annotation in the 3D domain. With the intrinsic calibration of the camera and its known position relative to the LiDAR, it is possible to project the annotations from the image onto the LiDAR points.

\subsubsection{Anonymisation}

The images were anonymised to ensure compliance with local regulations. This involved a thorough process where identifiable elements, such as number plates and faces, were identified and manually blurred to prevent potential privacy breaches. It was ensured that these elements were blurred beyond recognition and readability, and were double-checked to confirm. The anonymisation process was carried out to maintain the integrity of the visual data while adhering to legal and ethical protocols. 

\subsubsection{Annotation} We used the LabelMe \cite{labelmecitation} tool to annotate the images by outlining the road damage with polygons. The labels for the M2S-RoAD dataset capture nine key road damage, chosen for relevance to rural driving environments:

\textit{Pothole:}  Potholes are bowl-shaped depressions in the road surface, usually round or oval in shape. They are characterised by rough, jagged edges that are often surrounded by loose debris. Potholes can range from shallow to deep and are common on a variety of road surfaces, including asphalt, dirt, and grass.

\textit{Potholes with Water:}  Potholes appear very differently when filled with water, as reflections off the water distort camera and LiDAR readings; thus, potholes with water were classified separately than ordinary potholes. Water-filled potholes pose a hazard because the water masks the true depth and boundaries of the depression, making it difficult to accurately assess the hazard.

\textit{Cracks:}  Cracks are fractures in the pavement that appear as distinct lines either longitudinally or transversely. Cracks are typically narrower than slippage and may form around areas where potholes have been patched. Cracks appear both independently and in clusters, spanning large portions of the road.

\textit{Corrugation:} Also known as washboards, corrugations consist of a series of closely spaced ridges and valleys created in the road surface. This effect is common on unpaved roads, especially those made of dirt or gravel, and is caused by repeated traffic. 

\textit{Rutting:}  Rutting are longitudinal depressions formed along wheel paths of roads, caused by the repeated passage of heavy vehicles. Over time, road materials compress under the pressure of traffic, forming grooves that can lead to unsafe driving conditions.  It is most recognisable by the width of the tracks, matching the wheel width of cars.

\textit{Slippage:}  Slippage defects appear as wide, crescent-shaped cracks or areas where the top layer of pavement slips or separates from the underlying layers. Slippage is often seen at intersections or areas that experience frequent turning, sometimes including the shoulder or side of the road. Slipping is an early indicator of potential pavement failure and often occurs on ageing or poorly maintained asphalt roads.

\textit{Drain:} Drain covers themselves can pose a hazard in the road, and more pertinently, the road around or along a drain typically involves some kind of angle or channel to guide the water. These areas or channels are important for vehicles to recognise as they stray from the expected uniformity of the road and after rainy weather can also be prime locations for hydroplaning.

\textit{Bump:}  Bumps are raised areas in the pavement that can be caused by frost heave, underlying root growth, or construction defects, as well as intentional additions such as speed bumps. Bumps vary in size but can be differentiated from corrugation by frequency and compactness.

\textit{Manhole:}  Manholes are covered openings in the road, typically designed for maintenance use and providing access to underground. Additionally, poorly cared for manholes can mirror larger bumps or potholes in the road, and the road around manholes is particularly susceptible to more extreme damage.

\normalsize
\subsubsection{Point Cloud Projection}

Labelling road damage in the point cloud domain is not a straightforward process. We leverage the image annotations and sensor calibration to map the labels onto the point cloud data.

\begin{figure}[b!]
\centering

\begin{subfigure}[]{\columnwidth}
\centering
\includegraphics[width=\columnwidth]{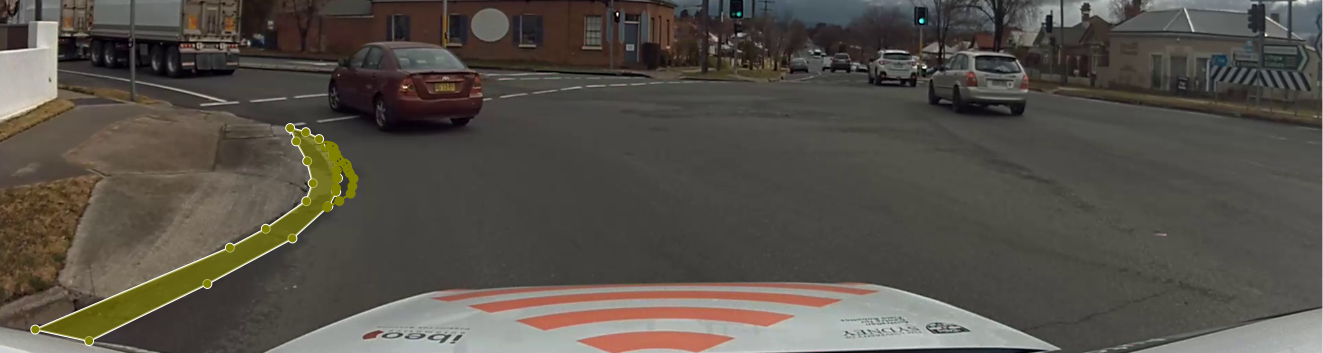}
\caption{\small Image with slippage annotation}
\label{fig:instance}
\end{subfigure}

\begin{subfigure}[]{\columnwidth}
\centering
\includegraphics[width=\columnwidth]{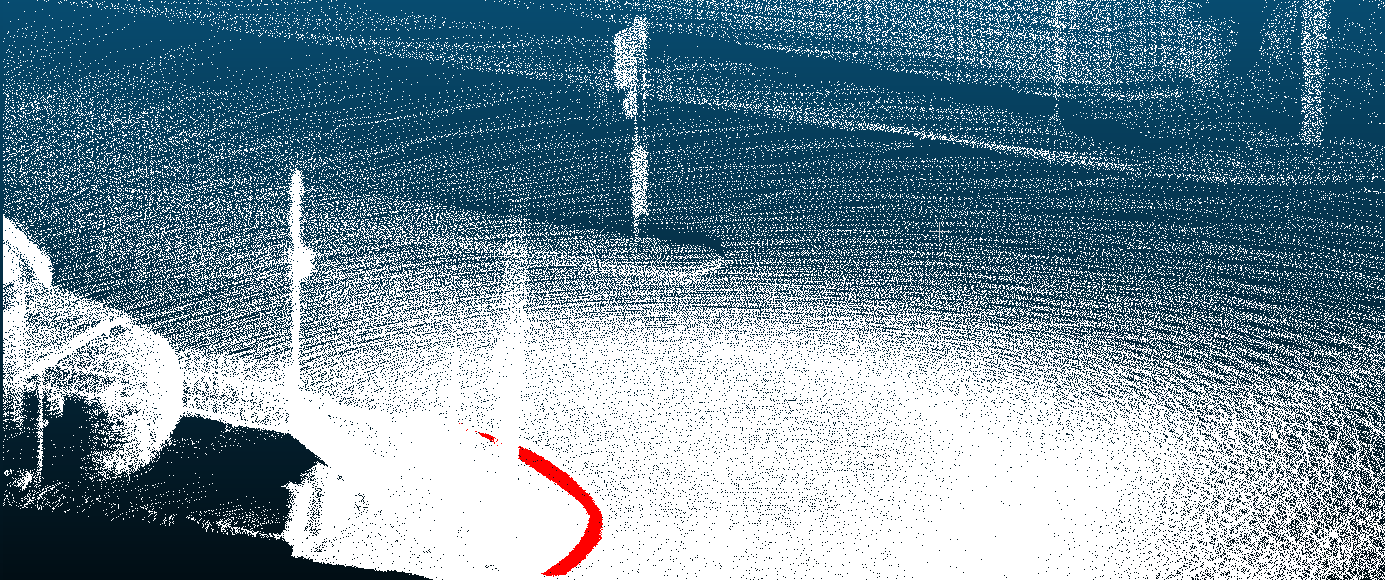}
\caption{\small Slippage label transferred to the point cloud}
\label{fig:panoptic}
\end{subfigure}
\caption{\small  Label transfer from image to point cloud domain.}
\label{fig:pc_projection}
\end{figure} 
Using the individual scans corresponding to the labelled images, we calculate the position of the LiDAR for each scan. To achieve this, we use KISS-ICP \cite{vizzo2023ral} to calculate these positions in space. We then concatenate all the LiDAR scans to build a denser point cloud. We project the points onto the image using the camera's intrinsic parameters and the extrinsic calibration between the two sensors. 

Once the 3D points are mapped to the image frame, we retrieve labels for the points that fall within the annotated polygons and assign these labels to the corresponding 3D points. Fig. \ref{fig:pc_projection} depicts the outcome of this process, with the right image displaying the annotated image and the left figure showing the projection of the labels onto the point cloud. The code for this transformation is available at \url{https://github.com/chinitaberrio/M2S-RoAD}.

\subsection{Dataset Statistics}

\begin{figure}[h!]
     \centering
     \includegraphics[width=\linewidth]{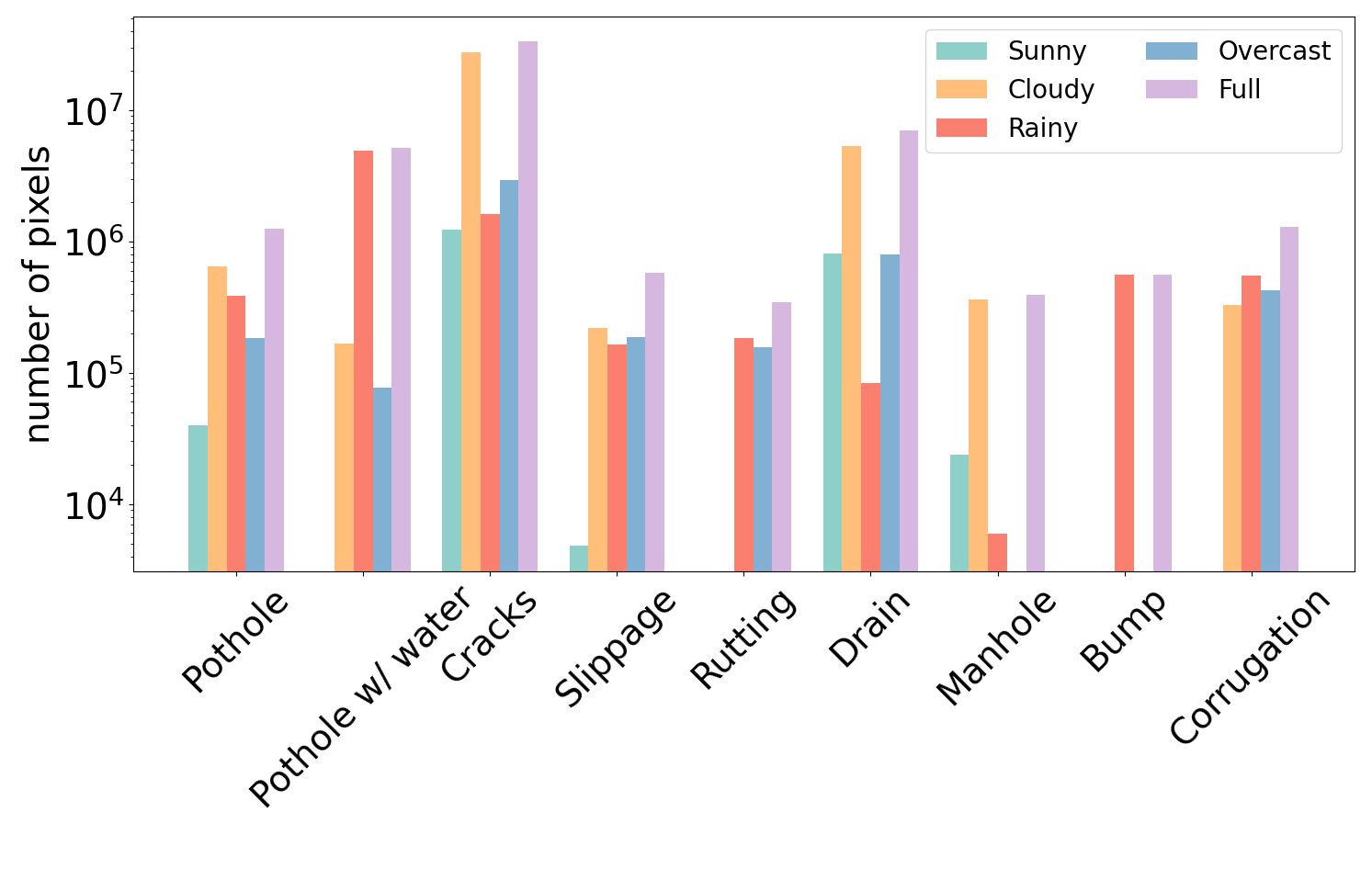}
     \caption{\small Distribution of annotated pixels (y-axis) across each class (x-axis) in the M2S-RoAD dataset and its subsets.}
     \label{fig:distribution-label}
 \end{figure}

The Fig. \ref{fig:distribution-label} depicts the distribution of road damage labels in the M2S-RoAD dataset and its subsets. The dataset subsets are categorised based on weather conditions with an additional category labelled ``full", which represents the combined distribution across all weather conditions.

Cracks are the most dominant label across the dataset, the `drain' label also shows a large presence in the \textit{sunny} subset. Pothole with water represents almost 60\% of the annotations for the \textit{rainy} dataset. Other damage types, such as potholes, slippage, manholes, bumps, and corrugations, are less common, which indicates their relative scarcity in the dataset.

\section{Semantic Segmentation Experiments}
In this section, we present semantic segmentation experiments on the M2S-RoAD dataset. We outline the task setup, evaluation metrics, and baseline methods. These methods were evaluated on the full validation set and subsets for four weather conditions, categorising results into five colour-coded sections. Finally, we discuss and analyse the outcomes to assess their performance.

\subsubsection{Task and Metrics}

Semantic segmentation aims to classify every pixel in an image, or each point in point clouds, enabling detailed 2D or 3D scene understanding. We utilise images with a resolution of 1928×1208 to generate segmentation maps that categorise nine types of road damage, including potholes, cracks, slippage, rutting, and manhole.

To assess a semantic segmentation model, we employ the mean Intersection over Union (mIoU) metric, sometimes referred as the mean Jaccard Index. This metric averages the Intersection over Union (IoU) \cite{everingham2015pascal} across all classes:

\[
\text{IoU}_i = \frac{TP_i}{TP_i + FP_i + FN_i}
\]

\[
\text{mIoU} = \frac{1}{N} \sum_{i=1}^{N} \text{IoU}_i
\]

where \( TP_i \), \( FP_i \), and \( FN_i \) denote the number of true positives, false positives, respectively, and false negatives for class \( i \), and \( N \) represents the total number of classes.

\subsubsection{Baseline Approaches}
We assessed five state-of-the-art semantic segmentation techniques on the M2S-RoAD dataset: ISANet \cite{huang2019interlaced}, BiSeNetV2 \cite{yu2021bisenet}, STDC \cite{fan2021rethinking}, SegFormer \cite{xie2021segformer}, and Mask2Former \cite{cheng2022masked}.

The BiSeNet series \cite{yu2018bisenet, yu2021bisenet} employs a dual-branch network architecture with a guided aggregation layer to integrate features from both branches. However, the additional stream reduces the computational efficiency.
In order to solve this problem, STDC \cite{fan2021rethinking} proposes a short-term dense connection network that  extracts image features with multi-scale information and expandable receptive fields. It incorporates spatial context into the low-level details, enhancing performance without lengthening the inference time.

ISANet \cite{huang2019interlaced} enhances efficiency by employing Interleaved Sparse Self-Attention (ISSA). ISSA divides the dense similarity matrix into two sparse matrices. Built upon ResNet \cite{he2016deep}, ISANet uses ISSA to generate segmentation outputs that align with the spatial dimensions of the input image.

Latest progress in semantic segmentation have been driven by transformer-based techniques. A framework called SegFormer \cite{xie2021segformer} integrates multi-layer perceptrons (MLP) with transformers. The MLP decoder then combines multi-scale features to create representations for producing segmentation maps.
Mask2Former \cite{cheng2022masked} introduces a Masked-attention Mask Transformer for semantic segmentation. Its decoder improves feature encoding and speeds up converging by concentrating on anticipated mask regions through masked attention. 

We adapted the implementation of baseline methods from the MMSegmentation library for training and evaluation on the M2S-RoAD dataset. BiSeNetV2 \cite{yu2021bisenet} was trained from scratch, while the backbones of the other four methods were initialised with ImageNet-1K \cite{deng2009imagenet} pre-trained weights.

\subsubsection{Results and Discussion}
\begin{table}[t!]
\centering
\caption{\small \textbf{Semantic segmentation results on the M2S-RoAD validation set and its subsets}. The results are organised by weather conditions: \raisebox{0pt}[0pt][0pt]{\colorbox{lightblue}{Blue}} indicates the \textit{sunny} subset (first section), \raisebox{0pt}[0pt][0pt]{\colorbox{lightyellow}{Yellow}} represents the \textit{cloudy} subset (second section), \raisebox{0pt}[0pt][0pt]{\colorbox{lightgray}{Gray}} shows the \textit{rainy} subset (third section), and \raisebox{0pt}[0pt][0pt]{\colorbox{lightred}{Red}} corresponds to the \textit{overcast} subset (fourth section). \raisebox{0pt}[0pt][0pt]{\colorbox{lightgreen}{Green}} highlights results for the entire validation set (fifth section). The evaluated methods include ISANet \cite{huang2019interlaced}, BiSeNetV2 \cite{yu2021bisenet}, STDC \cite{fan2021rethinking}, SegFormer \cite{xie2021segformer}, and Mask2Former \cite{cheng2022masked}. A hyphen (-) indicates unavailable or inapplicable data.}
\resizebox{\columnwidth}{!}{%
\begin{tabular}{l|c|ccccccccc}
\toprule
\textbf{Method} & \rotatebox{90}{\textbf{mIoU}} & \rotatebox{90}{\textbf{pothole}} & \rotatebox{90}{\parbox{2cm}{\textbf{pothole w/ water}}} & \rotatebox{90}{\textbf{cracks}} & \rotatebox{90}{\textbf{slippage}} & \rotatebox{90}{\textbf{rutting}} & \rotatebox{90}{\textbf{drain}} & \rotatebox{90}{\textbf{manhole}} & \rotatebox{90}{\textbf{bump}} & \rotatebox{90}{\textbf{corrugation}} \\
\midrule
\rowcolor{lightblue} BiSeNetV2 & 36.65 & 0.43 & - & 89.58 & 0.0 & - & 93.23 & 0.0 & - & - \\
\rowcolor{lightblue} STDC & 60.87 & 21.35 & - & 91.14 & 0.0 & - & 91.84 & 100.0 & - & - \\
\rowcolor{lightblue} ISANet & 71.49 & 70.20 & - & 93.41 & 0.0 & - & 93.82 & 100.0 & - & - \\
\rowcolor{lightblue} SegFormer & 65.72 & 45.9 & - & 91.87 & 0.0 & - & 90.83 & 100.0 & - & - \\ 
\rowcolor{lightblue} Mask2Former & 59.08 & 40.27 & - & 81.62 & 0.0 & - & 73.7 & 99.79 & - & - \\
\midrule
\rowcolor{lightyellow} BiSeNetV2 & 45.87 & 12.08 & 32.21 & 95.8 & 0.0 & - & 87.42 & 47.7 & - & - \\
\rowcolor{lightyellow} STDC & 68.93 & 49.61 & 54.39 & 96.44 & 46.12 & - & 87.97 & 79.05 & - & - \\
\rowcolor{lightyellow} ISANet & 80.88 & 56.62 & 63.59 & 96.91 & 99.96 & - & 89.2 & 79.0 & - & - \\
\rowcolor{lightyellow} SegFormer & 62.39 & 51.64 & 63.4 & 97.39 & 60.05 & - & 90.74 & 73.53 & - & - \\
\rowcolor{lightyellow} Mask2Former & 67.33 & 41.93 & 62.93 & 96.13 & 32.7 & - & 93.26 & 77.03 & - & - \\
\midrule
\rowcolor{lightgray} BiSeNetV2 & 20.99 & 15.12 & 74.55 & 31.79 & 0.0 & 0.0 & - & - & 0.0 & 25.48 \\
\rowcolor{lightgray} STDC & 52.71 & 16.47 & 91.12 & 45.84 & 21.37 & 44.99 & - & - & 97.88 & 51.31 \\
\rowcolor{lightgray} ISANet & 50.75 & 12.96 & 87.37 & 39.32 & 22.87 & 48.02 & - & - & 96.33 & 48.4 \\
\rowcolor{lightgray} SegFormer & 53.23 & 15.02 & 86.59 & 47.87 & 23.73 & 45.97 & - & - & 97.49 & 55.96 \\
\rowcolor{lightgray} Mask2Former & 49.36 & 24.68 & 84.72 & 30.95 & 21.03 & 45.03 & - & - & 96.82 & 42.3 \\
\midrule
\rowcolor{lightred} BiSeNetV2 & 53.64 & 41.74 & 55.83 & 84.14 & 0.0 & - & 89.69 & - & - & 50.44 \\
\rowcolor{lightred} STDC & 64.23 & 72.16 & 75.4 & 89.15 & 5.19 & - & 96.87 & - & - & 46.6 \\
\rowcolor{lightred} ISANet & 60.86 & 77.9 & 53.89 & 89.71 & 0.0 & - & 95.28 & - & - & 48.37 \\
\rowcolor{lightred} SegFormer & 66.50 & 62.2 & 57.49 & 93.05 & 17.82 & - & 97.73 & - & - & 70.73 \\
\rowcolor{lightred} Mask2Former & 74.92 & 58.22 & 69.51 & 89.77 & 75.48 & - & 93.04 & - & - & 63.47 \\
\midrule
\rowcolor{lightgreen} BiSeNetV2 & 38.65 & 15.76 & 73.26 & 89.99 & 0.0 & 0.0 & 87.64 & 43.24 & 0.0 & 37.97 \\
\rowcolor{lightgreen} STDC & 65.26 & 36.31 & 89.1 & 93.39 & 20.66 & 33.0 & 89.57 & 77.02 & 97.9 & 50.4 \\
\rowcolor{lightgreen} ISANet & 65.47 & 33.47 & 85.86 & 93.09 & 26.17 & 36.68 & 90.14 & 80.03 & 95.97 & 47.82 \\
\rowcolor{lightgreen} SegFormer & 67.06 & 33.48 & 85.62 & 94.0 & 26.65 & 34.79 & 91.38 & 74.74 & 97.49 & 65.41 \\
\rowcolor{lightgreen} Mask2Former & 67.77 & 33.5 & 83.45 & 91.45 & 37.25 & 44.23 & 90.97 & 78.14 & 96.81 & 54.12 \\
\bottomrule
\end{tabular}%
}
\label{tab:results_semantic}
\end{table}

Table \ref{tab:results_semantic} shows semantic segmentation results for the M2S-RoAD dataset, highlighting performance differences across four weather conditions and the nine road damage types. The methods achieve their best overall average performance of 65.08\% on the \textit{cloudy} subset, while their lowest performance is 45.41\% on the \textit{rainy} subset. The reason for this difference is that cloudy weather provides ample light and visibility, along with 34 million labelled pixels for diverse training samples. In contrast, rainy conditions face image quality challenge, like, lens fogging, raindrops, and reflections. 

The performance of the baseline methods is closely linked to the representativeness of road damage types in the dataset. For instance, when segmenting cracks and rutting, the models achieve average performances of 92.38\% and 29.74\%, respectively. This disparity arises because cracks account for 66.67\% of labelled pixels, making them the most representative samples, while rutting represents only 0.69\%. This imbalance enables effective feature learning for cracks but limits support for rutting, ultimately impacting overall performance.

The weather-specific categories show significantly better performance on relevant weather subsets. For example, when segmenting potholes with water, the baseline methods achieve an average performance of 84.87\% in the \textit{rainy} subset, compared to just 55.30\% and 62.42\% in the \textit{cloudy} and \textit{overcast} subsets, respectively. This disparity likely stems from the fact that potholes with water primarily occur in rainy conditions, making up 58.11\% of labelled pixels in the \textit{rainy} subset, while representing less than 2\% in the other two subsets. 

Transformer-based methods outperform CNN-based approaches in road damage segmentation. Mask2Former achieves the highest performance at 67.77\%, while BiSeNetV2 has the lowest performance at only 38.65\%. This gap is largely due to the larger parameter count in Transformers and their global self-attention mechanism, which better captures features and details of different road damage.

\section{Conclusion}

This paper introduces a new dataset for semantic segmentation of diverse road surface damage types, and illustrates the diverse environmental and weather-related challenges faced by road surfaces in New South Wales, Australia.

From the experiment analysis, we discovered that some classes need more images to be labelled. In the future, we will focus on expanding the dataset to make it more balanced and improve the performance of machine learning methods.

Although we successfully transferred the labels from the image to the point cloud, we will evaluate the semantic segmentation task for 3D points in the future.

This work serves as the starting point for multi-modal segmentation of road damage. We aim to develop a method combining lidar and camera data to detect and segment road damage.

\bibliographystyle{apalike}
\bibliography{acra}

\end{document}